\pdfoutput=1

\documentclass[11pt]{article}

\usepackage[]{ACL2023}

\usepackage{times}
\usepackage{latexsym}

\usepackage[T1]{fontenc}
\usepackage{CJKutf8}

\usepackage[utf8]{inputenc}

\usepackage{microtype}

\usepackage{inconsolata}

\usepackage{graphicx}
\graphicspath{ {./} }

\usepackage{soul}
\usepackage{xcolor}
\newcommand{\ctext}[3][RGB]{%
  \begingroup
  \definecolor{hlcolor}{#1}{#2}\sethlcolor{hlcolor}%
  \hl{#3}%
  \endgroup
}

\usepackage{booktabs}
\usepackage{stfloats}
\title{Kanbun-LM: Reading and Translating Classical Chinese\\in Japanese Methods by Language Models}

\author{Hao Wang \quad Hirofumi Shimizu \quad Daisuke Kawahara \\
        Waseda University \\
         \texttt{{\{conan1024hao@akane., bowen1205@toki., dkw@\}waseda.jp}} \\}
\begin{document}
\begin{CJK}{UTF8}{ipxm}
\maketitle
\begin{abstract}
Recent studies in natural language processing (NLP) have focused on modern languages and achieved state-of-the-art results in many tasks.
Meanwhile, little attention has been paid to ancient texts and related tasks.  
Classical Chinese first came to Japan approximately 2,000 years ago.
It was gradually adapted to a Japanese form called Kanbun-Kundoku (Kanbun) in Japanese reading and translating methods, which has significantly impacted Japanese literature.
However, compared to the rich resources for ancient texts in mainland China, Kanbun resources remain scarce in Japan.
To solve this problem, we construct the first Classical-Chinese-to-Kanbun dataset in the world.
Furthermore, we introduce two tasks, character reordering and machine translation, both of which play a significant role in Kanbun comprehension.
We also test the current language models on these tasks and discuss the best evaluation method by comparing the results with human scores.
We release our code and dataset on GitHub\footnote{https://github.com/nlp-waseda/Kanbun-LM}.
\end{abstract}
\section{Introduction}
Classical Chinese was introduced to Japan approximately 2,000 years ago~\citep{japanese-history}.
Then Classical Chinese began to be adapted to a Japanese form in Japanese reading and translating methods in the 8th century A.D.~\citep{kanbun-asia}.
This form is called \textit{Kanbun-Kundoku}.
For simplicity, we call it Kanbun in this paper.
Kanbun has influenced many famous Japanese literary works, such as \textit{Manyoshu}~\citep{manyoshu} and \textit{The Tale of Genji}~\citep{genji-2}.
To this day, Kanbun still occupies 50 points out of 200 in the common test for Japanese university admissions, which shows the deep influence of Kanbun on Japanese culture.

\begin{figure}[t]
    \centering
    \includegraphics[width=0.75\linewidth]{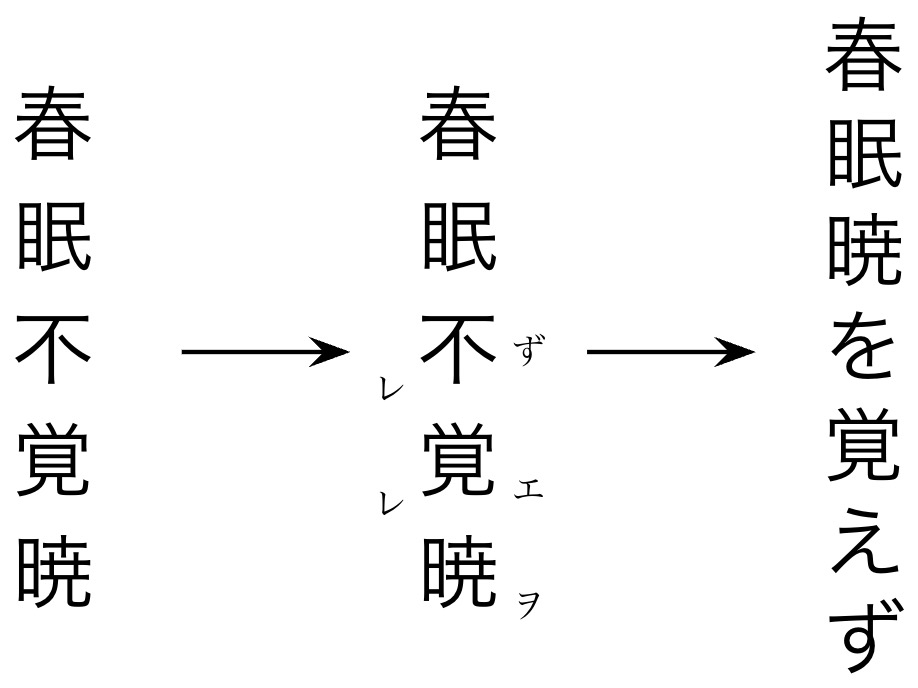}
    \caption{An example of Kanbun. ``春眠不覚暁'' (en: This morning of spring in bed I’m lying) is original Classical Chinese. To transform it into Kanbun, we first add Kaeriten, Yomigana, and Okurigana to the sentence. Two ``レ'' on the left side are Kaeriten, indicating the characters need to be read in reverse. On the right side, there is a Yomigana ``ず'', meaning ``不'' should be written as ``ず''. ``エ'' and ``ヲ'' are Okurigana, making the sentence from isolating to agglutinative. Now if we read the sentence following the above rules, the sentence becomes ``春眠暁を覚えず'' (While adding marks, we use Katakana like ``エ'' and ``ヲ'', but in a complete sentence, we use Hiragana like ``え'' and ``を''. They have no difference except for their looks).}
    \label{fig:ex-kanbun}
\end{figure}

Although Chinese and Japanese have many characters in common, reading Classical Chinese is not easy for Japanese people because of the following two reasons.
First, Chinese (also Classical Chinese) is in SVO (Subject-Verb-Object) word order, which is the same as English.
On the other hand, Japanese is in SOV (Subject-Object-Verb) word order, which leads to difficulties in understanding Chinese.
Second, Chinese is an isolating language with little to no morphological variation and a nearly one-to-one ratio of morphemes to words. However, Japanese is an agglutinative language that attaches prefixes and suffixes to a word to indicate the grammatical relationship of that word in a sentence.
These differences led to the creation of Kanbun.
To make the text from SVO to SOV, from isolating to agglutinative, Japanese people developed a system of various conventional reading punctuation, diacritical and syntactic markers~\citep{crawcour1965introduction}.
We list the three main types of markers below and show a specific example of Kanbun in Figure~\ref{fig:ex-kanbun}.
Since the Kanbun system is highly sophisticated, we omit to explain all the rules in this paper.
There are also other systems for reading Classical Chinese in other regions like Korean Peninsula~\citep{korean} and Khitan, but we focus on the Japanese Kanbun system in this paper.

\paragraph{Kaeriten}
(ja:返り点) marks placed on the left side of characters indicating the characters need to be read in reverse, making the sentence from SVO to SOV. (e.g., ``我有レ兄'' (en:I have a brother) should be read as ``我兄有'', ``レ'' is the mark)
\paragraph{Yomigana}
(ja:読み仮名) Hiragana (Japanese phonological units) that are placed on the right side of characters, indicating the characters' reading in Japanese. (e.g., ``不'' (en:no) is read as ``ず'')
\paragraph{Okurigana}
(ja:送り仮名) Katakana (Phonological units, collectively referred to as Kana with Hiragana) that are placed on the right side of characters for making the sentence from isolating to agglutinative. (e.g., the Chinese character ``飲'' (en:drink) is ``飲む'' in Japanese, which has an extra Kana)
\\

Compared to the vast amount of research and language resources available for Classical Chinese, there is little research on Kanbun, and the language resources for Kanbun are highly scarce.
For instance, over 48,900 Tang poems (poems written in the characteristic style of the Tang dynasty) are included in \textit{Quan Tangshi} and are all accessible via the Internet.
However, to our knowledge, only around 500 Tang poems adapted to Kanbun are accessible.
This large gap makes the research on Kanbun increasingly difficult.
Although a lot of data of Kanbun exists in ancient books, it is beyond our ability to apply OCR to them and compile the results into clean data.
Therefore, building a high-performance Classical-Chinese-to-Kanbun translator is the most efficient way to address the lack of Kanbun language resources.
Moreover, understanding the mechanisms of Kanbun will also lead to understanding Classical Japanese literature (such as Wakan konkōbun, a mixture of Japanese and Chinese writing styles), as well as Japanese culture and thought.

In previous work, \citet{ud-kaeriten, ud-tree, ud-ud} proposed a series of applications for Classical Chinese using Universal Dependencies~\citep{nivre-etal-2016-universal}.
\citet{ud-kundoku, ud-kundoku-2} proposed a method for Classical-Chinese-to-Kanbun machine translation.
However, this method is rule-based and less precise, and the author did not make a dataset to conduct a quantitative evaluation.
In this work, we construct the first Classical-Chinese-to-Kanbun dataset in the world.
Based on this, we introduce Kanbun-LM, where we fine-tune language models for reading and translating Classical Chinese in Japanese methods, trying to fill the resource gap.

The main contributions of our work are summarized as follows:
\begin{itemize}
    \item We construct the first Classical-Chinese-to-Kanbun dataset in the world, which addresses the lack of Kanbun language resources.
    \item We introduce two tasks for the dataset, character reordering and machine translation, both of which are significant in Kanbun comprehension. We conduct quantitative evaluations for both tasks and achieved state-of-the-art results in both tasks using language models, which has shown major improvement over the baseline~\citep{ud-kundoku, ud-kundoku-2}. We also construct a pipeline for the tasks and verify whether pre-reordering is helpful to machine translation.
    \item We discuss the best evaluation method for Classical-Chinese-to-Kanbun translation by comparing the results with human scores, which is not covered in existing work.
\end{itemize}
\section{Related Work}
\subsection{Work for Classical Chinese}
Although Classical Chinese is still an unstudied field, it has enough resources for exploration compared to other low-resource ancient texts.
\textit{Daizhige}\footnote{https://github.com/garychowcmu/daizhigev20} contains approximately 3.3 billion tokens and is the largest dataset for Classical Chinese.
The \textit{Siku Quanshu} corpus is made from the largest collection of books in Chinese history, with 36,381 volumes and approximately 997 million words.
Chinese-Poetry\footnote{https://github.com/chinese-poetry/chinese-poetry} is a database that contains more than 300,000 ancient Chinese poems.
There are also several corpora with extra information that can be used for downstream tasks.
For example, the Ancient Chinese Corpus (ACC)\footnote{https://catalog.ldc.upenn.edu/docs/LDC2017T14} is a dataset of \textit{Zuo Zhuan} (a Pre-Qin Chinese book published late in the 4th century BC) that contains the information of word segmentation and POS tags.

Since BERT~\citep{devlin-etal-2019-bert} and BERT-like models~\citep{liu2019roberta, https://doi.org/10.48550/arxiv.1909.11942, he2021debertav3} were proposed, pre-training language models on a large corpus and fine-tuning them on downstream tasks have become a paradigm in NLP studies.
In the Classical Chinese field, several pre-trained models have also been proposed.
SikuBERT and SikuRoBERTa~\citep{sikubert} are pre-trained on the \textit{Siku Quanshu} corpus and evaluated on the following four tasks using the ACC dataset: word segmentation, punctuation restoration, POS tagging, and named entity recognition.
GuwenBERT\footnote{https://github.com/ethan-yt/guwenbert} is pre-trained on the \textit{Daizhige} corpus and evaluated on the CCLUE\footnote{https://cclue.top} benchmark.
Meanwhile, GPT~\citep{radford2019language}-based models such as SikuGPT2\footnote{https://huggingface.co/JeffreyLau/SikuGPT2} and T5~\citep{2020t5}-based models such as Mengzi-T5~\citep{https://doi.org/10.48550/arxiv.2110.06696} are also proposed for text generation.

To evaluate the general performance of pre-trained language models, benchmarks for natural language understanding (NLU) tasks have been proposed in many languages.
For Classical Chinese, CCLUE provides five NLU tasks, including sentence segmentation, named entity recognition, text classification, and text retrieval.
Recently, WYWEB~\citep{wyweb} has been proposed. It contains eight tasks, including sentence classification, sequence labeling, reading comprehension, and machine translation.
\subsection{Work for Kanbun}
\citet{ud-kaeriten} proposed a method to reorder Classical Chinese sentences to Japanese reading order using dependency parsing by Universal Dependencies~\citep{nivre-etal-2016-universal}.
First, the method applies morphological analysis to Classical Chinese sentences to segment them into tokens and assign POS tags.
Second, it obtains dependency relations using the arc-planar algorithm~\citep{gomez-rodriguez-nivre-2010-transition}, which was mainly trained on Universal Dependencies of \textit{Mengzi}, \textit{Lunyu}, and \textit{Liji} (these are all ancient Chinese books).
Finally, it applies character reordering based on the results of dependency parsing and 24 rules proposed by the author.

Furthermore, \citet{ud-kundoku, ud-kundoku-2} proposed an encode-reorder-decode model, called UD-Kundoku, to translate Classical Chinese to Kanbun, while the encoding and reordering modules take the approaches introduced in \citet{ud-kaeriten}.
To make the reordered sentences into Kanbun, the author introduced a rule-based decoding module that adds Okurigana to sentences and makes the sentences from isolating to agglutinative.
Okurigana can be roughly divided into two categories: auxiliary words and inflectional suffixes.
The rules also support special characters, such as characters left unpronounced and characters that need to be read twice when reading Kanbun.

\citet{ud-kundoku-2} also conducted a brief evaluation for generated Kanbun results using BLEU~\citep{papineni-etal-2002-bleu} and RIBES~\citep{ribes}.
However, the author only evaluated a few examples and did not make an in-depth discussion.
\section{Our Dataset and Tasks}
We construct a parallel dataset for Classical Chinese and Kanbun.
The dataset consists of original ancient Chinese texts, Japanese reading orders, and Kanbun texts.
We show examples in Table~\ref{tab:data-example}.

Although it is crucial to choose texts that cover as many periods as possible since vocabulary and grammar change with time, it is difficult to construct a comprehensive dataset.
To our knowledge, \textit{Tangshixuan}\footnote{https://kanbun.info/syubu/toushisen000.html} (Selection of Tang Poems) is the largest resource containing both original ancient Chinese texts and translated Kanbun texts.
We use this resource to make our dataset.
For pre-processing, we extract the Japanese reading order from Kanbun by a rule-based program.
For the special tokens that may not appear in Kanbun or appear multiple times, we annotated them manually.
We also convert the characters from old character forms to new character forms (kind of like transforming Traditional Chinese to Simplified Chinese, but in Japanese character forms) using dictionaries to mitigate the out-of-vocabulary problem.

\textit{Tangshixuan} contains a total of 465 poems.
We split the dataset using group shuffle split to ensure that all sentences in one poem would not be split.
Table~\ref{tab:statistics} lists the statistics of the dataset.

Based on the dataset, we introduce two tasks, character reordering and machine translation, both of which are significant in Kanbun comprehension.
For character reordering, the goal is to transform Classical Chinese texts into Japanese reading orders, from SVO to SOV.
Japanese reading orders as shown in Table~\ref{tab:data-example}, such as ``12543'', are the targets to be predicted.
Machine translation is a sequence-to-sequence task that translates Classical Chinese texts into Kanbun.
Since the source and target sentences share the vocabulary, it can also be considered as a multilingual rewriting task.

\begin{figure*}[t]
    \centering
    \includegraphics[width=1\linewidth]{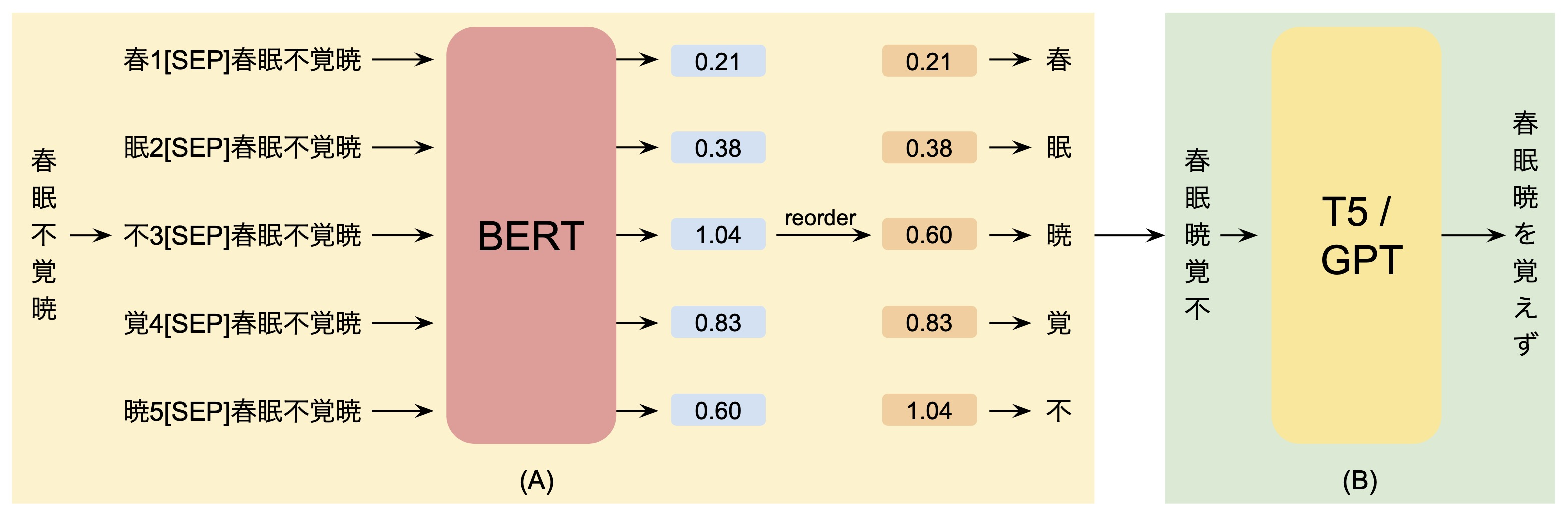}
    \caption{An overview of the pipeline. (A) is the character reordering module and (B) is the machine translation module. (A) receives original Classical Chinese sentences and reorders them into Japanese reading order. (B) receives reordered sentences from (A) and translates them into Kanbun.}
    \label{fig:pipeline}
\end{figure*}

\begin{table*}[t]
\centering
\resizebox{2\columnwidth}{!}{%
\begin{tabular}{ccll}
\hline
Classical Chinese & Japanese reading order & Kanbun & (English tr.)\\
\hline
春眠不覚暁 & 12543 & 春眠暁を覚えず & This morning of spring in bed I’m lying\\
処処聞啼鳥 & 12453 & 処処啼鳥を聞く & Not wake up till I hear birds crying\\
夜来風雨声 & 12345 & 夜来風雨の声 & After one night of wind and showers\\
花落知多少 & 12345 & 花落つること知んぬ多少ぞ & How many are the fallen flowers\\
\hline
\end{tabular}}
\caption{Examples of our dataset. Each instance has a triple of the original ancient Chinese text, the Japanese reading order (the numbers represent their index in the original text), and the translated Kanbun text.}
\label{tab:data-example}
\end{table*}

\begin{table}[t]
\centering
\resizebox{0.85\columnwidth}{!}{%
\begin{tabular}{lrrr}
\hline
Split & Poems & Sentences & Characters\\
\hline
Train & 372 & 2,731 & 16,411 \\
Validation & 46 & 320 & 2,038 \\
Test & 47 & 370 & 2,254 \\
\hline
\end{tabular}}
\caption{Statistics of our dataset. The number of characters refers to the original ancient Chinese data.}
\label{tab:statistics}
\end{table}
\section{Experimental Setup}
\subsection{Implementation for Tasks}
In this section, we introduce our implementation details of the two tasks: character reordering and machine translation.
We also construct a pipeline for the two tasks and verify whether pre-reordering is helpful to machine translation.
We use NVIDIA A100 (40GB) for the experiments.
Figure~\ref{fig:pipeline} shows an overview of our pipeline.

For character reordering, we propose a rank-based sorting method that fine-tunes BERT-like models to predict the rank (position in Japanese reading order) for every character in a sentence.
We split each sentence into characters and preprocess them into inputs by the form \ctext[RGB]{220,220,220}{\{character\}\{the character's index in the sentence\}[SEP]\{sentence\}}.
The character's index is added to handle the cases where more than two identical characters appear in one sentence.
To make gold labels for training, we normalize the ranks by the lengths of the sentences, making the value of ranks range from 0 to 1 (for a sentence of length 5, the ranks will be normalized from 1, 2, ..., 5 to 0.2, 0.4, ..., 1).
Once we collect the output ranks, we sort them in ascending order and restore them to the original characters.
Then we obtain a reordered sentence.
An illustration of our sorting method is shown in (A) of Figure~\ref{fig:pipeline}.

For machine translation, we simply fine-tune T5 and GPT to generate Kanbun from original Classical Chinese sentences.
Since we want to see the real level of each model, we did not apply any filter to the generations.

For the pipeline, we pass original Classical Chinese sentences to the character reordering module first, making them from SVO to SOV.
Then we pass the sorted sentences to the machine translation module to add Okurigana, transforming from isolating to agglutinative.
\subsection{Pre-trained Models}
\subsubsection{Models for Character Reordering}
We conduct experiments on five models in total for character reordering.
Two models are pre-trained on Japanese corpora, two on Chinese corpora, and one on Classical Chinese corpora.
All of the models' tokenizers are character-based because we intend to predict the exact position of each character.
We do not use multilingual models like mBERT~\citep{devlin-etal-2019-bert} and XLM-RoBERTa~\citep{conneau-etal-2020-unsupervised} because their tokenizers do not generally expect character-based encoding.\footnote{mBERT can tokenize Chinese into characters effectively. However, there is no guarantee that it tokenizes Japanese into characters too, since not all Japanese characters are in the CJK Unicode range.}
We use the following five models, all in base size, consisting of 12 layers, 768 dimensions of hidden states, and 12 attention heads.
We show more details of the models in Appendix~\ref{appendix:model} and details of fine-tuning hyper-parameters in Appendix~\ref{appendix:parameter}.
\paragraph{BERT-japanese-char}
This model is trained on the Japanese version of Wikipedia.
\paragraph{RoBERTa-japanese-char-wwm}
This model is trained on the Japanese version of Wikipedia and the Japanese portion of CC-100~\citep{conneau-etal-2020-unsupervised}.
The whole word masking (wwm)~\citep{Cui_2021} strategy is applied.
\paragraph{BERT-chinese}
This model is trained on the Chinese version of Wikipedia.
\paragraph{RoBERTa-chinese-wwm-ext}
This model is trained on 5.4B tokens, which include the Chinese version of Wikipedia and extra data. 
The whole word masking strategy is applied.
\paragraph{RoBERTa-classical-chinese-char}
This model is derived from GuwenBERT.
Simplified characters' embeddings are expanded to traditional characters, making vocabulary size larger.
\subsubsection{Models for Machine Translation}
We use mT5~\citep{xue-etal-2021-mt5} and mGPT~\citep{https://doi.org/10.48550/arxiv.2204.07580} for machine translation experiments. 
We do not use Japanese models because the vocabulary size is much smaller than multilingual models, and they generate many [UNK] tokens, leading to unreadable generations.
We show more details of the models in Appendix~\ref{appendix:model} and details of fine-tuning hyper-parameters in Appendix~\ref{appendix:parameter}.
\paragraph{mT5}
mT5 is trained on the mC4~\citep{2020t5} corpus, covering 101 languages (Chinese and Japanese are both contained).
We use small, base, and large models in our experiments.
\paragraph{mGPT}
This model is trained on 60 languages using Wikipedia and the mC4 corpus (Chinese and Japanese are both contained).
\subsection{Automatic Evaluation Metrics}
\subsubsection{Metrics for Character Reordering}
Following the previous sentence reordering studies~\citep{cui-etal-2020-bert, Kumar_Brahma_Karnick_Rai_2020, https://doi.org/10.48550/arxiv.2103.13584}, we use the following metrics for evaluation.
\paragraph{Kendall's Tau ($\tau$)}
This metric measures the rank correlation between two sentences.
Fewer the number of inversions needed to sort predicted character orders into ground truth character orders means stronger correlation and better performance.
$$ \tau = 1 - \frac{4(\#inversions)}{\#char(\#char-1)}$$
\paragraph{Perfect Match Ratio (PMR)}
This metric measures the percentage of predicted character orders exactly matching with ground truth orders.
\subsubsection{Metrics for Machine Translation}
There is no systematic work on evaluating Classical-Chinese-to-Kanbun translation.
On top of BLEU and RIBES, which are used by \citet{ud-kundoku-2}, we add ROUGE~\citep{lin-2004-rouge} and BERTScore~\citep{bert-score} for our experiments, trying to maintain the diversity of evaluation metrics.
We implemented all these metrics on the basis of characters since word-based evaluation highly depends on morphological analysis, and related packages for Kanbun are still immature.
\paragraph{BLEU}
BLEU~\citep{papineni-etal-2002-bleu} is the most widely used metric in machine translation.
It is an n-gram-based metric that computes the exact match precision scores of n-grams that occur in the reference and the candidate.
\paragraph{RIBES}
RIBES~\citep{ribes} is a rank-based metric proposed to evaluate machine translation between languages with widely differing word orders.
It applies word mapping to the reference and the candidate first, and then computes rank correlation as scores for the evaluation.
\paragraph{ROUGE}
ROUGE~\citep{lin-2004-rouge} is a commonly used n-gram-based metric for summarization evaluation.
\citet{lin-2004-rouge} proposed ROUGE-n, which computes the exact match recall scores of n-grams, and ROUGE-L, which computes scores using longest common subsequence instead.
Since ROUGE-1, ROUGE-2, and ROUGE-L did not show much difference in our experiments, 
we only report ROUGE-L's results in this paper.
\paragraph{BERTScore}
BERTScore~\citep{bert-score} is an embedding-based metric that computes a similarity score for each token in the candidate with each token in the reference.
To calculate character-based scores, we use BERT-japanese-char (layer 11) in our experiments.
\subsection{Manual Annotations}
We recruited three people who are bilingual in Chinese and Japanese as our human annotators.
There are two criteria for annotator selection: (1) ability to read Classical Chinese in original word order; (2) ability to get full marks in the Kanbun part of the Japanese university admission exam.

For character reordering, to compare with the models, we asked the annotators to do the same sorting task, which the models did, with no access to reference materials and the Internet.
We collected results, computed Kendall's Tau and PMR scores, and averaged them.

For machine translation, we asked the annotators to evaluate models' generations according to the following three metrics, rated on a 5-point scale from 1 to 5 (larger is better).
The reference sentences were also evaluated to measure the quality of the dataset.
The annotators were allowed to search for reference materials in this evaluation.
\paragraph{Relevance}
This rating measures how well the translation is done, which judges
whether the content is translated without any shortage or deviation.
\paragraph{Accuracy}
This rating measures the quality of a generation, which judges
whether it is lexically and grammatically correct in Japanese.
\paragraph{Fluency}
This rating measures the fluency and naturalness of a generation and whether the rhythm of Classical Chinese remains.
\section{Results and Discussion}
\subsection{Character Reordering}
The results of the character reordering task are presented in Table~\ref{tab:reorder}.
UD-Kundoku is the baseline method that was proposed by \citet{ud-kundoku, ud-kundoku-2}.
Human scores are the average of the three annotators' results.

All the BERT-like models outperformed the baseline and human scores.
The two Chinese models performed slightly better than the two Japanese models, and RoBERTa-classical-chinese-char, which was pre-trained on the ancient Chinese corpus, performed the best.
Compared to the baseline, RoBERTa-classical-chinese-char achieved 22.5\% better Kendall's Tau and 94.7\% better PMR scores.
Compared to human scores, RoBERTa-classical-chinese-char achieved 11.8\% better Kendall's Tau and 29.2\% better PMR scores.

\paragraph{Gap between the Chinese and Japanese models.}
Since more ancient texts are present in a Chinese corpus like Wikipedia, we speculate that the score gap between the Chinese and Japanese models originates from the pre-training corpus rather than the reading orders of the pre-training languages.
Considering that this task requires converting SVO to SOV, it would be ideal to use both Chinese and Japanese corpora for pre-training.
However, since the existing multilingual models cannot guarantee to tokenize an input text into characters, we leave this validation to future work.

\paragraph{Additional data did not help.}
The two RoBERTa models did not score higher than the two BERT models. This is probably because many ancient texts do not exist in the additional corpus like CC-100~\citep{conneau-etal-2020-unsupervised}, and thus the additional training in RoBERTa did not strengthen the models' understanding of Classical Chinese.

\paragraph{BERT is more accurate in details.}
When comparing with human scores, we had an interesting finding that although the PMR scores of humans and RoBERTa-japanese-char-wwm are similar, Kendall's Tau score of the model is 5.9\% higher.
This indicates that BERT is more accurate than humans in predicting the details of the orders.
Although our annotators are bilingual, they are not experts in Classical Chinese.
We hope to collaborate with real experts in the future to conduct experiments and see if BERT can still retain an advantage.

\paragraph{Error analysis.}
Since the PMR score of our best model is 0.783, most predicted orders are exactly correct. 
However, we still found some error patterns that the model encountered.
It is not easy to distinguish whether a pair of two characters is a noun or a combination of a verb and a noun.
Moreover, determining the order becomes challenging when two verbs appear in a sentence.

\begin{table}[t]
\centering
\resizebox{\columnwidth}{!}{%
\begin{tabular}{lcc}
\hline
Model Setup & $\tau$ & PMR\\
\hline
UD-Kundoku & 0.770 & 0.402 \\
Human & 0.844 & 0.606 \\
\hline
BERT-japanese-char & 0.898 & 0.637 \\
RoBERTa-japanese-char-wwm & 0.894 & 0.600 \\
BERT-chinese & 0.917 & 0.689 \\
RoBERTa-chinese-wwm-ext & 0.920 & 0.718 \\
RoBERTa-classical-chinese-char & \textbf{0.944} & \textbf{0.783} \\
\hline
\end{tabular}}
\caption{Kendall's Tau ($\tau$) and PMR scores of character reordering. UD-Kundoku is the baseline, and human scores are the average of the three annotators' results.}
\label{tab:reorder}
\end{table}
\subsection{Machine Translation}
\begin{table*}[t]
\centering
\resizebox{1.8\columnwidth}{!}{%
\begin{tabular}{lccccccc}
\hline
Model Setup & BLEU & RIBES & ROUGE-L & BERTScore & Relevance & Accuracy & Fluency \\
\hline
UD-Kundoku & 0.097 & 0.309 & 0.546 & 0.884 & - & - & - \\
reference & - & - & - & - & 4.958 & 4.951 & 4.949 \\
\hline
mT5-small & 0.317 & 0.428 & 0.659 & 0.914 & 3.219 & 3.002 & 3.153 \\
mT5-base & 0.462 & 0.520 & 0.735 & 0.930 & - & - & - \\
mT5-large & \textbf{0.514} & \textbf{0.583} & \textbf{0.747} & \textbf{0.934} & \textbf{3.948} & \textbf{3.884} & \textbf{3.904} \\
mGPT & 0.303 & 0.476 & 0.606 & 0.898 & 2.548 & 2.270 & 2.236 \\
\hline
\end{tabular}}
\caption{Results of machine translation, containing the automatic and manual evaluation metrics. UD-Kundoku is the baseline, and reference is the Kanbun target of translation.}
\label{tab:translation}
\end{table*}

\begin{table*}[t]
\centering
\resizebox{2\columnwidth}{!}{%
\begin{tabular}{llll}
\hline
Model Setup & (a) & (b) & (c) \\
\hline
input & 投筆事戎軒 & 駆馬出関門 & 鳳林戈未息 \\
reference & 筆を投じて戎軒を事とす & 馬を駆って関門を出づ & 鳳林戈未だ息まず \\
\hline
mT5-small & 筆を投じて戎軒を事す & 馬を駆って関門に出づ & 鳳林戈未だ息し \\
mT5-base &  筆を投じて戎軒に事す & 馬を駆って関門に出で & 鳳林戈未だ息まず \\
mT5-large & 筆を投じて戎軒を事とす & 馬を駆って関門を出づ & 鳳林戈未だ息まず \\
mGPT & 筆を投じて戎軒に事とすを事 & 馬を駆って関門を出でんとすも出で & 鳳林戈未だ息まずかとすかとす鳳 \\
\hline
(English tr.) & Laid down my pen and turned to the war & Mounted my horse and left through the gates & The forest's battle drums remain unabated \\
\hline
\end{tabular}}
\caption{Generation examples of machine translation. Input is the original Classical Chinese sentence, and reference is the Kanbun target of translation.}
\label{tab:gen_example}
\end{table*}

\subsubsection{Model Performance}
Table~\ref{tab:translation} lists the results of machine translation, which contains the automatic and manual evaluation metrics.
UD-Kundoku is the baseline, and the reference is the Kanbun target.

For the automatic evaluation, all our models exceeded the baseline in all evaluation metrics.
The performance of mT5 increased as the model size increases, with mT5-large performing best.
The performance of mGPT and mT5-small are close to each other.

For the human evaluation, we asked annotators to evaluate only the translations of mT5-small, mT5-large, and mGPT. This is because mT5-base performs close to mT5-large, and the baseline's results are too poor to be evaluated.
As with the automatic evaluation, mT5-large performed the best.
On the other hand, mT5-small significantly outperformed mGPT in this evaluation.
The reference sentences obtained very high scores, proving that our dataset's Kanbun data is of high quality.
We also calculated Fleiss' Kappa to measure Inter-Annotator Agreement (IAA).
The results of Fleiss' Kappa for relevance, accuracy, and fluency are 0.360, 0.371, and 0.341, which show fair agreements~\citep{kappa}.

\paragraph{Generation examples.}
We show three generation examples in Table~\ref{tab:gen_example}.
In all three examples, mT5-large performed flawlessly, giving the same translations as the reference.
mT5-base and mT5-small generated translations similar to mT5-large, but with some minor errors.
mGPT sometimes repeated the characters in the original sentences (``事'' in (a), ``出'' in (b), and ``鳳'' in (c)), which lowers the scores of human evaluation.
``未'' in (c) is an example of special characters that need to be read twice, which should be read as ``未だ...ず'' (en:yet).
In this case, mT5-base and mT5-large generated the correct translation.
However, mT5-small and mGPT could not recognize it as a special character.

\paragraph{Why is mGPT so weak?}
Although mGPT has almost 1.5 times the number of parameters of mT5-large (detailed model sizes can be found in Appendix~\ref{appendix:model}), its translations are not even as good as mT5-small.
Since mT5 and mGPT are both mainly trained on mC4~\citep{2020t5}, the effect of the pre-training corpus can be largely excluded.
One reason is the repetition of words that we have explained before.
For other reasons, we speculate that the encoder modules in mT5 have a significant role in comprehending Classical Chinese.
However, this is only a hypothesis and needs to be tested with more future experiments.
\subsubsection{Correlation between Evaluation Metrics}
We show Pearson and Spearman correlation coefficients between the automatic evaluation metrics and human evaluation metrics in Table~\ref{tab:correlation}.
BERTScore has the greatest correlation with all three human evaluation metrics.
BLEU and ROUGE-L also performed well.
The rank-based metric, RIBES, performed the worst.
We notice that, compared to BLEU and ROUGE-L, BERTScore only has a slight lead in the correlation with relevance.
However, the advantage has increased in correlation with accuracy and fluency.
We speculate that this is because BERTScore can potentially capture sequence information~\citep{bert-score}, which makes it more possible to judge whether a sentence is accurate and fluent.
We also speculate that BERTScore better suits Classical-Chinese-to-Kanbun because Kanbun is generally very short, which can cause BLEU and ROUGE to be influenced by small changes.

We also show the correlation between the human evaluation metrics in Table~\ref{tab:correlation}.
Accuracy and fluency have the greatest correlation, which indicates that grammatically and lexically correct sentences are also fluent.
In general, the correlation between the metrics is relatively high.
To consider more different perspectives, we hope to reduce the correlation by discussing with Classical Chinese experts and reformulating the manual evaluation metrics in future work.

\begin{table}[t]
\centering
\resizebox{1\columnwidth}{!}{%
\begin{tabular}{lcccccc}
\midrule
Metric & \multicolumn{2}{c}{Relevance} & \multicolumn{2}{c}{Accuracy} & \multicolumn{2}{c}{Fluency} \\
\cmidrule(rr){2-3} \cmidrule(rr){4-5} \cmidrule(rr){6-7}
 & $r$ & $\rho$ & $r$ & $\rho$ & $r$ & $\rho$ \\
\midrule
BLEU & 0.667 & 0.650 & 0.637 & 0.605 & 0.594 & 0.576 \\
RIBES & 0.480 & 0.497 & 0.453 & 0.449 & 0.389 & 0.417 \\
ROUGE-L & 0.688 & 0.677 & 0.631 & 0.610 & 0.599 & 0.584 \\
BERTScore & \textbf{0.707} & \textbf{0.691} & \textbf{0.671} & \textbf{0.642} & \textbf{0.644} & \textbf{0.625} \\
\midrule
Relevance & - & - & 0.862 & 0.849 & 0.835 & 0.829 \\
Accuracy & 0.862 & 0.849 & - & - & \textbf{0.946} & \textbf{0.947} \\
Fluency & 0.835 & 0.829 & \textbf{0.946} & \textbf{0.947} & - & - \\
\midrule
\end{tabular}}
\caption{Pearson ($r$) and Spearman ($\rho$) correlation coefficients for relevance, accuracy, and fluency between automatic metrics and human judgment. We also show the correlation between each human evaluation metric.}
\label{tab:correlation}
\end{table}
\subsection{Pipeline}
\begin{table}[t]
\centering
\resizebox{\columnwidth}{!}{%
\begin{tabular}{lcccc}
\hline
Model Setup & BLEU & RIBES & ROUGE-L & BERTScore\\
\hline
mT5-small & 0.317 & 0.428 & 0.659 & 0.914 \\
 + reorder & 0.328 & 0.420 & 0.701 & 0.916 \\
 + reorder (gold) & \textbf{0.359} & \textbf{0.451} & \textbf{0.727} & \textbf{0.919} \\
\hline
mT5-base & \textbf{0.462} & 0.520 & 0.735 & 0.930\\
 + reorder & 0.413 & 0.486 & 0.735 & 0.926 \\
 + reorder (gold) & 0.461 & \textbf{0.529} & \textbf{0.770} & \textbf{0.932} \\
\hline
mT5-large & \textbf{0.514} & \textbf{0.583} & 0.747 & 0.934\\
 + reorder & 0.479 & 0.551 & 0.748 & 0.931 \\
 + reorder (gold) & 0.502 & 0.573 & \textbf{0.774} & \textbf{0.935} \\
\hline
mGPT & 0.303 & 0.476 & 0.606 & 0.898 \\
 + reorder & 0.303 & 0.467 & 0.612 & 0.894 \\
 + reorder (gold) & \textbf{0.340} & \textbf{0.508} & \textbf{0.642} & \textbf{0.900} \\
\hline
\end{tabular}}
\caption{Results of the pipeline. The first row of each model is the direct end-to-end machine translation results. The second row (``+ reorder'') uses RoBERTa to sort characters before doing machine translation. The third row (``+ reorder (gold)'') does pre-reorder by gold labels instead of RoBERTa's predictions.}
\label{tab:pipeline}
\end{table}
We show the pipeline results in Table~\ref{tab:pipeline}.
The first row of each model is the direct machine translation results, which are also shown in Table~\ref{tab:translation}.
The second row (``+ reorder'') shows the results using RoBERTa-classical-chinese-char to reorder characters before passing the sentences to machine translation.
The third row (``+ reorder (gold)'') uses the gold labels of the reading orders instead of the predictions by RoBERTa to reorder characters.

By pre-reordering using RoBERTa, most of the evaluation metrics of mT5-small were improved.
mGPT basically remained at the original level.
While mT5-base and mT5-large showed a decreasing trend in most of the metrics.
We speculate that as the model's performance increases, the model will gradually be able to do character reordering and machine translation at the same time.
Since the predictions of RoBERTa are not 100\% accurate, wrong predictions may confuse models and lead to their inability to determine correct orders.

In contrast, by pre-reordering using the gold labels, all models received some degree of improvement in almost all evaluation metrics.
This indicates that correct pre-reordering does help machine translation, and it is necessary to do more work on improving the character reordering module.
\section{Conclusion and Future Work}
In this paper, to address the lack of Kanbun language resources, we used language models to read Classical Chinese in Japanese reading orders and translate Classical Chinese into Kanbun.
We constructed the first Classical-Chinese-to-Kanbun dataset in the world, which includes original ancient Chinese texts, translated Kanbun texts, and the Japanese reading orders.

Furthermore, we introduced two tasks for the dataset: character reordering and machine translation.
We achieved state-of-the-art results in both tasks, which have a great lead over the baseline.
We also constructed a pipeline for the two tasks and verified that accurate pre-reordering is helpful for machine translation.
However, the accuracy of current reordering models is not enough, and future efforts are needed to improve the accuracy.

Moreover, we discussed which automatic evaluation metric is the most suitable for Classical-Chinese-to-Kanbun translation by computing the correlation between the automatic and human evaluation metrics.
In our experiments, BERTScore is the best.
However, we only tested with character-based metrics.
More experiments are still needed to test subword-based and sentence-based metrics.

In the future, we hope to continuously update the dataset to include an increasingly comprehensive range of ancient texts.
We also hope to collaborate with experts in Classical Chinese to find the upper bound of human character reordering accuracy, refine the manual evaluation metrics to a more streamlined one, and make a deeper exploration on the best automatic evaluation metric.
\section*{Limitations}
Due to the lack of data, our dataset is not comprehensive since it only consists of Tang poems.
Our model may not perform well on unseen data in other forms.
We plan to update the dataset in the future continuously.

Our evaluation metrics and generation results for the machine translation tasks are not certified by experts in Classical Chinese, so the results and discussions in this paper are not entirely reliable.
We welcome more experts and researchers to join our work in the future.

Due to the limitation of GPU resources, we do not experiment on larger models.
We welcome researchers to test our method on large models and make some deeper discussions. 
\section*{Acknowledgements}
This work was supported by JSPS KAKENHI Grant Number JP21H04901.
We are grateful to the annotators who have spent much of their time helping with the experiments. 
We would also like to thank the reviewers for their insightful comments for improving the paper.
\bibliography{anthology,custom}
\bibliographystyle{acl_natbib}
\appendix
\onecolumn
\section{Details of pre-trained models}
\label{appendix:model}

We show the details of the pre-trained models used in our experiments below.
Table~\ref{tab:model1} lists the details of the BERT-like models for character reordering, and Table~\ref{tab:model2} lists those of the pre-trained models for machine translation.

\begin{table*}[h]
\centering
\caption{Details of pre-trained models (character reordering).}
\resizebox{\columnwidth}{!}{%
\begin{tabular}{llrrrr}
\hline
model & corpus & \#dimension & \#layers & \#heads & vocabulary size \\
\hline
BERT-japanese-char & Wikipedia (ja) & 768 & 12 & 12 & 6,144 \\
\footnotesize{(cl-tohoku/bert-base-japanese-char-v2)} & & & & & \\
RoBERTa-japanese-char-wwm & Wikipedia (ja) + CC-100 (ja) & 768 & 12 & 12 & 18,377 \\
\footnotesize{(ku-nlp/roberta-base-japanese-char-wwm)} & & & & & \\
BERT-chinese & Wikipedia (zh) & 768 & 12 & 12 & 21,128 \\
\footnotesize{(bert-base-chinese)} & & & & & \\
RoBERTa-chinese-wwm-ext & Wikipedia (zh) + ext & 768 & 12 & 12 & 21,128 \\
\footnotesize{(hfl/chinese-roberta-wwm-ext)} & & & & & \\
RoBERTa-classical-chinese-char & Wikipedia (zh) + \textit{Daizhige} + ext & 768 & 12 & 12 & 26,318 \\
\footnotesize{(KoichiYasuoka/roberta-classical-chinese-base-char)} & & & & & \\
\hline
\end{tabular}}
\label{tab:model1}
\end{table*}

\begin{table*}[h]
\centering
\caption{Details of pre-trained models (machine translation).}
\resizebox{0.8\columnwidth}{!}{%
\begin{tabular}{llrrrrr}
\hline
model & corpus & \#params & \#dimension & \#layers & \#heads & vocabulary size \\
\hline
mT5-small & mC4 (101 languages) & 172M & 512 & 8 & 6 & 250,112 \\
\footnotesize{(google/mt5-small)} & & & & & & \\
mT5-base & mC4 (101 languages) & 390M & 768 & 12 & 12 & 250,112 \\
\footnotesize{(google/mt5-base)} & & & & & & \\
mT5-large & mC4 (101 languages) & 973M & 1024 & 24 & 16 & 250,112 \\
\footnotesize{(google/mt5-large)} & & & & & & \\
mGPT & Wikipedia + mC4 & 1,417M & 2048 & 24 & 16 & 100,000 \\
\footnotesize{(sberbank-ai/mGPT)} & (both 60 languages) & & & & & \\
\hline
\end{tabular}}
\label{tab:model2}
\end{table*}
\section{Hyper-parameters}
\label{appendix:parameter}
\label{sec:appendix}

We show the hyper-parameters used in our experiments in Table~\ref{tab:hyperparamters}.
The numbers in the curly brackets indicate that grid searches were performed to select the best fit.

\begin{table*}[h]
\centering
\caption{Hyper-parameters used in the experiments.}
\resizebox{0.6\columnwidth}{!}{%
\begin{tabular}{lc}
\hline
hyper-parameter & value\\
\hline
learning rate & \{1e-5, 2e-5, 5e-5\} \\
batch size & \{8, 16, 32\} \\
epoch & \{1-20\} (BERT), \{10, 20, 30\} (T5), \{1, 2, 3\} (GPT) \\
\hline
\end{tabular}}
\label{tab:hyperparamters}
\end{table*}
\end{CJK}
\end{document}